\documentclass{article}

\usepackage{microtype}
\usepackage{graphicx}
\usepackage{subfigure}
\usepackage{booktabs} 

\usepackage{hyperref}

 \usepackage[accepted]{icml2023}

\usepackage{amsmath}
\usepackage{amssymb}
\usepackage{mathtools}
\usepackage{amsthm}

\usepackage[capitalize,noabbrev]{cleveref}

\theoremstyle{plain}

\theoremstyle{definition}

\theoremstyle{remark}

\usepackage[textsize=tiny]{todonotes}

\usepackage{graphicx} 
\usepackage{array}
\usepackage{amsmath}
\usepackage{amssymb}
\usepackage{graphicx}
\usepackage{caption}
\usepackage{cleveref}
\usepackage{balance}

\newcommand{\papertitle}{Detecting Dataset Drift and Non-IID Sampling via k-Nearest Neighbors}

\icmltitlerunning{\papertitle}

\begin{document}

\twocolumn[
\icmltitle{\papertitle}
\icmlsetsymbol{equal}{*}

\begin{icmlauthorlist}
\icmlauthor{Jesse Cummings}{comp,uni}
\icmlauthor{Elías Snorrason}{comp}
\icmlauthor{Jonas Mueller}{comp}
\end{icmlauthorlist}

\icmlaffiliation{uni}{MIT}
\icmlaffiliation{comp}{Cleanlab}

\icmlcorrespondingauthor{Jesse Cummings}{jecummin@mit.edu}
\icmlcorrespondingauthor{Jonas Mueller}{jonas@cleanlab.ai}

\icmlkeywords{Machine Learning, ICML}
\vskip 0.3in
]

\printAffiliationsAndNotice{}  

\begin{abstract}
We present a straightforward statistical test to detect certain violations of the assumption that the data are Independent and Identically Distributed (IID). The specific form of violation considered is common across real-world applications: whether the examples are ordered in the dataset such that almost adjacent examples tend to have more similar feature values (e.g.\ due to distributional drift, or attractive interactions between datapoints). Based on a k-Nearest Neighbors estimate, our approach can be used to audit any multivariate numeric data as well as other data types (image, text, audio, etc.) that can be numerically represented, perhaps via model embeddings.
Compared with existing methods to detect drift or auto-correlation, our approach is both applicable to more types of data and also able to detect a wider variety of IID violations in practice.

\end{abstract}

\section{Introduction}

Most machine learning and statistics relies on the foundational  assumption that the data is sampled in an IID manner. To reliably generalize to the population underlying the data, one assumes each datapoint does not affect the others (independent) and that the population is not changing as datapoints are being collected (identically distributed). In settings that violate these assumptions, data-driven inference cannot be trusted unless done with special care \cite{federated, decentralized, beyond}.

Real-world data collection can be messy and hard to know whether data was sampled in a strictly IID fashion. Unlike other works from the time-series and online learning literature, this paper considers a single given dataset (without an explicit time column) and presents a statistical test to answer a practical question: \emph{Does the order in which my data were collected matter}? The role of data-ordering/ collection-times is often not obvious, especially for non-expert data analysts. 
Algorithms that can automatically detect when data violates the IID assumption in certain ways are especially valuable to:  novice data scientists, users of AutoML systems, or those who simply lack domain knowledge (or time to acquire it) about their data.

Because it is \emph{impossible} to detect all the ways a dataset can be non-IID, an effective audit should aim to detect the types of violations most common in data-driven applications. In particular, such common violations include: \emph{drift} where the underlying distribution/population is evolving over time \cite{drift-def} and attractive interactions between certain datapoints which influence their values to be mutually similar \cite{point-processes, self-exciting}. Both types of common IID violations lead to data that exhibits the following \textbf{common non-IID property}: datapoints that are closer together in the data ordering tend to have more similar feature values.  

Although shuffling the order of a dataset is an effective way to make it appear IID, it remains important to determine if the underlying data collection or generation process is actually IID. IID violations may imply a dataset is not fully representative of the population and future data could look different – issues that cannot be addressed merely by shuffling the datapoints’ order. If the data distribution drifts over time in a ML application, then the most recently collected data will better represent future deployment conditions. In this case, holding-out this latest data is a more reliable way to estimate model performance vs. shuffling the dataset and using a random train/test split. This is just one example out of many where special actions are preferred once one realizes the data are not IID \cite{stanford-split,noniid-ml,quagmire}. Hence why an automated check for common violations of IID sampling is valuable, especially if it is simple and efficient!

Here, we introduce a straightforward k-Nearest Neighbors (kNN) approach\footnote{Code to run our method: \url{https://github.com/cleanlab/cleanlab} \\
Code to reproduce our benchmarks: \url{https://github.com/cleanlab/examples/blob/master/non_iid_detection/non_iid_detection.ipynb} } for detecting when data exhibits the \emph{common non-IID property}. Our approach performs statistical test of this alternative against the null hypothesis that the collection order of the datapoints does not affect their joint feature value distribution. The only requirement to apply our method is defining a similarity measure between two datapoints' feature values (to form a kNN graph over the data), and thus the method can be directly applied to complex multivariate data including images, text, audio, etc. (e.g.\ via cosine distance between model embeddings of each example). As empirically demonstrated in subsequent experiments, our kNN method is capable of detecting many forms of IID violation including datasets where the underlying distribution is: drifting as data are being collected, subject to a sharp changepoint, or marginally identical over time but the datapoints are not independent with positively-correlated interactions.

\section{Related Work}
With the growth in real-world model deployments, \emph{drift detection} has become increasingly relevant in recent years.  
Most new methods are intended for online use where a dataset is continually updated over time \cite{driftreview}. 
Offline methods like ours can nonetheless be easily integrated into an online application. 
Many diverse strategies for drift detection have been proposed, including methods based on: distributional dissimilarity \cite{ddm, rddm, ewddm}, examining datapoints sequentially to track change \cite{page, dmddm}, window-based examination of a datastream in batches \cite{adwin, ocdd, igrnn}, statistical pattern tracking \cite{nonparam, dynamic}, density clustering \cite{clustering}, and model-based interpretability \cite{explanation}. Each method balances trade-offs between sensitivity, generality, and complexity \cite{driftreview} and are at times not obviously extendable to multi-dimensional data. 

A kNN-based approach has been previously considered for \emph{concept drift detection} in data labels \cite{knn-based}, whereas our work is instead focused on the features of the data. 
More similar to our work are kNN-based approaches to \emph{two-sample testing} \cite{knn-tests, crossmatch}, which utilize nearest neighbor graphs to perform distribution-free two sample testing in arbitrary dimensions---a powerful framework that inspired our work. To our knowledge, none of these existing kNN-based statistical tests has been applied to a \emph{single} dataset in order to identify the types of IID violations considered in this paper.

\begin{figure*}[tb]
    \centering
    \subfigure[IID data]{\label{fig:gm_iid}\includegraphics[width=70mm]{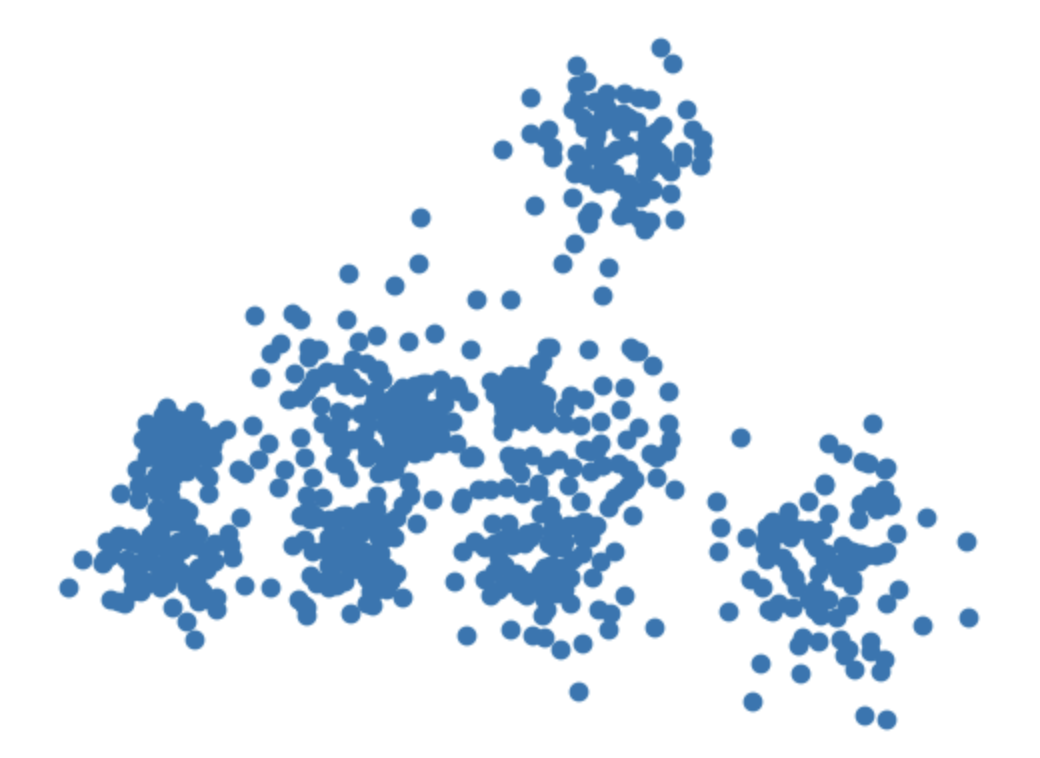}}
    \subfigure[Gradual mean shift]{\label{fig:gm_shift}\includegraphics[width=70mm]{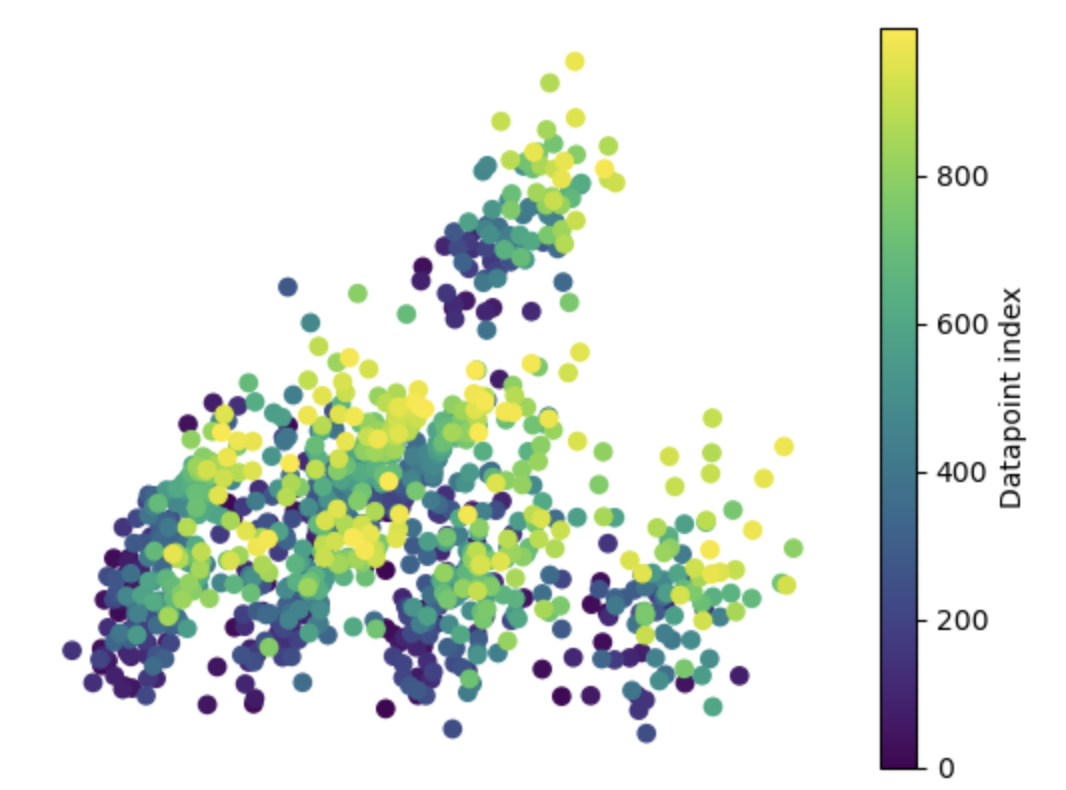}}
    \subfigure[Variance changepoint]{\label{fig:gm_changepoint}\includegraphics[width=70mm]{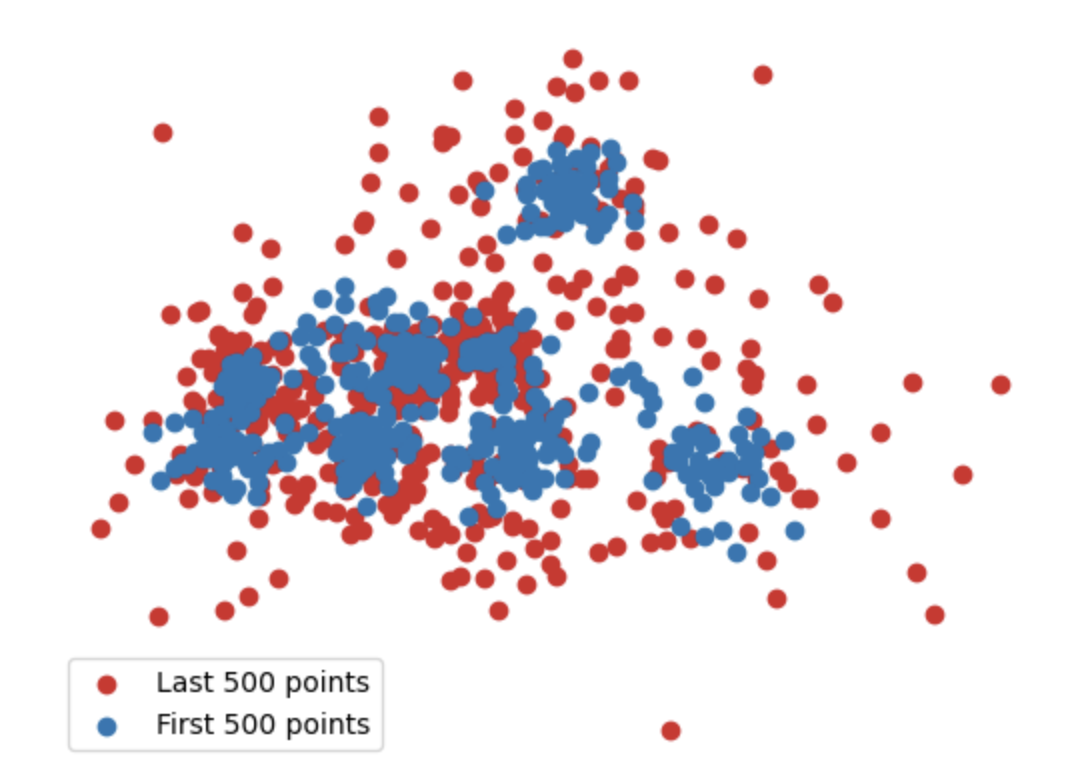}}
    \subfigure[Dependent but marginally identical]{\label{fig:dependent}\includegraphics[width=70mm]{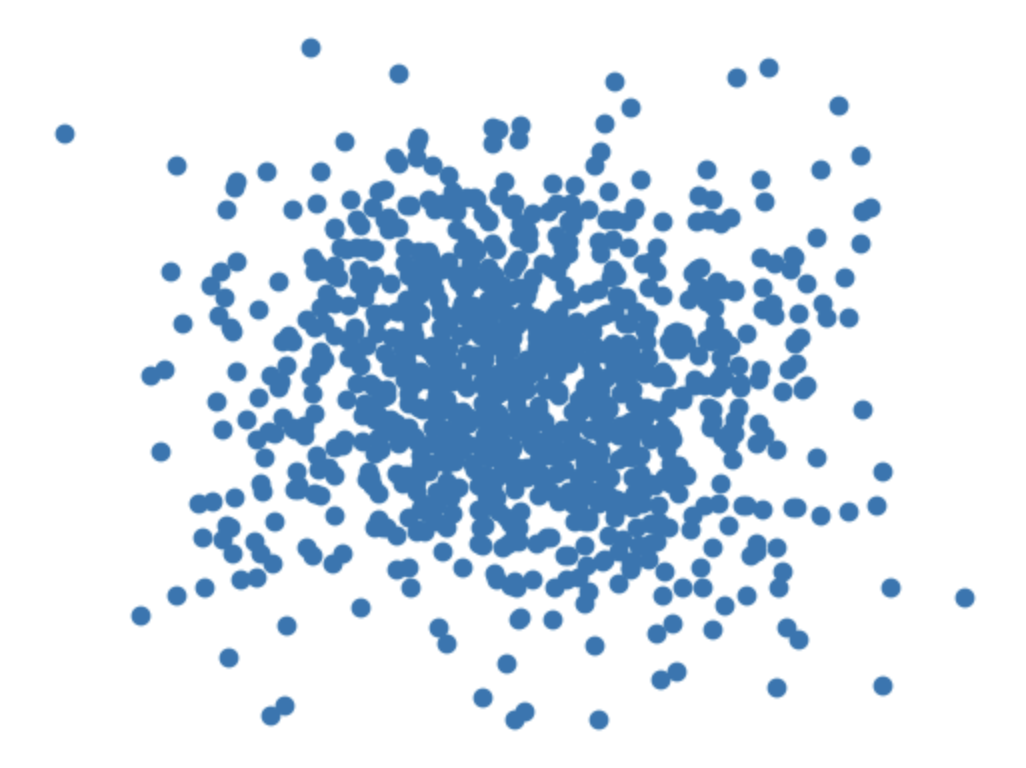}}
    \caption{Visualizations of the 2D datasets used for our benchmark. \textbf{(a-c)} Data is drawn from the same mixture of Gaussians. \textbf{(a)} Data is drawn IID from the distribution. \textbf{(b)} The mean of each Gaussian mixture component drifts gradually as each sample is drawn. In the plot, the color gradient represents the index of each sample in the dataset. \textbf{(c)} The variance of each Gaussian mixture component doubles suddenly after half the data is collected. In the plot, the colors indicate samples collected in the first vs.\ second halves of the dataset. \textbf{(d)} Samples are drawn in auto-regressive matter where their values are strongly inter-dependent but marginally identical in distribution.}
    \label{fig:toydata}
\end{figure*}

\begin{figure*}[tb]
    \centering
    \includegraphics[scale=0.5]{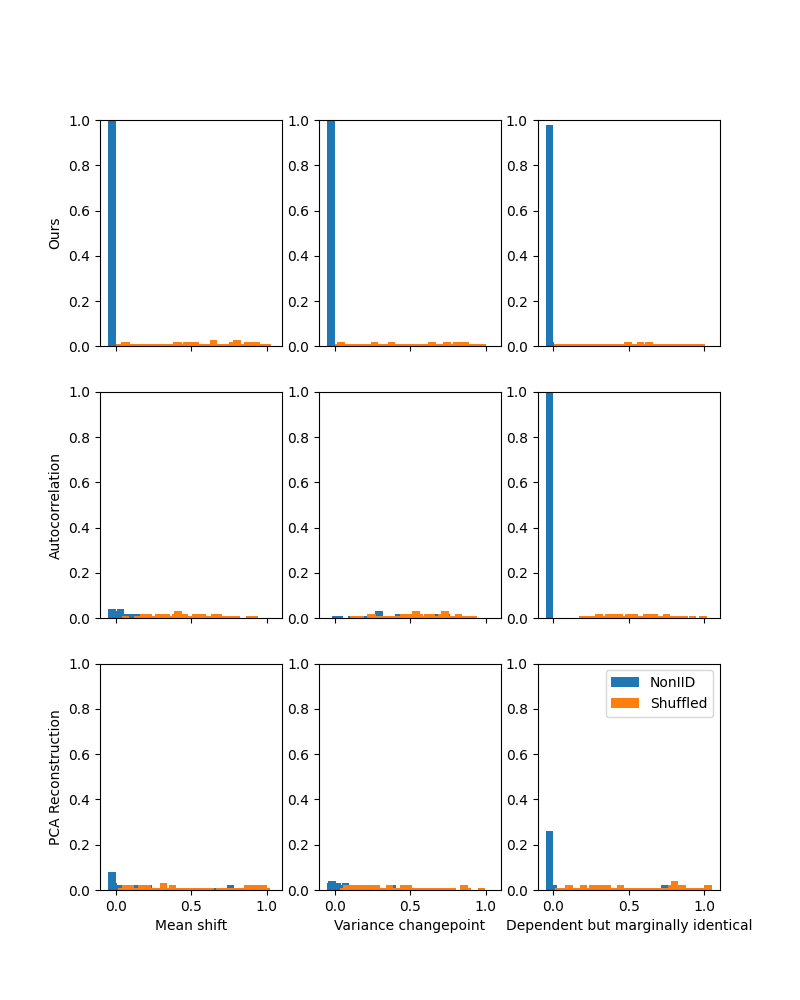}
    \vspace*{-2em}
    \caption{The distribution of p-values for different baselines on the Gaussian data scenarios described. Shown are histograms of p-values computed from 50 replicate runs of each method in each scenario. We also shuffled the data from each scenario to create an analogous IID dataset on which all methods were also run. Ours is the only method which reliably distinguishes between IID and non-IID data in each scenario.}
    \vspace*{2em}
    \label{fig:baselines}
\end{figure*}

\section{Methods}
Our method is simple yet effective. Given a similarity measure between datapoints' feature values (e.g.\ cosine distance for multivariate numerical data or embedding vectors), we construct a k-Nearest Neighbor graph of the dataset based on feature values. A pair of datapoints is considered \emph{neighbors} or \emph{non-neighbors} based on this kNN graph (\emph{not}  their indices in the dataset ordering). 

Next we consider the indices of the datapoints (perhaps determined by the time each was collected), and compute the \emph{index distance} between datapoint $i$ and $j$ as $|i - j|$ (e.g.\ the index distance between the 8th datapoint and the 108th datapoint is 100). Subsequently, we apply a two-sample permutation test using the Kolmogorov-Smirnov statistic in order to determine if there is a statistically significant difference between the distributions of index distances between kNN-neighbors (the \emph{foreground} distribution) vs. index distances between arbitrary datapoint pairs (the \emph{background} distribution). A low p-value implies that neighbors and non-neighbors in feature space are systematically ordered in certain patterns, indicating the data was sampled in a non-IID manner. At a high-level, this is all there is to our straightforward kNN approach, but we provide some lower-level details for completeness.

\subsection{Dataset Statistic}
Given a dataset $D = \{x_1, x_2, ..., x_N\}$ containing $N$ examples $x_i$, we first collect the index distances of all pairs of neighbors in the
dataset. Let $$X = \{ |i - j| \;| \; \forall x_i \in D, \; j \in K_i \}.$$

Where $K_i$ is the set of $k$ indices corresponding to the $k$ neighbors of
  $x_i$. For example, if $k = 2$ and $x_i$ is neighbors with $x_{i +
    1}, x_{j}$ then $K_i = \{i + 1, j\}$.
Throughout, we stick with the default value of $k=10$ and did not empirically observe the choice of $k$ to have much impact as long as $k$ was not too large. 
    
Next, we define the background distribution, or the distribution of
all possible index distances. We use a cumulative distribution to represent the probability of randomly drawing a pair of points from dataset $D$ with index distance less than or equal to $d$. This is easily defined analytically as a function mapping an index-distance $d$ to the empirical distribution function (over random pairs of datapoints) evaluated at that value: 
$$B(d) = \sum_{d' = 1}^d \frac{2 (N - d')}{N (N - 1)}.$$

As there are $\frac{N(N - 1)}{2}$ total pairs, $(N - d')$ of which which have index distance exactly $d'$. 

The test statistic $T$ for our dataset is computed as follows:
$$T = \max_{d \in \{1,2,..., N-1\}} |\widehat{F}_X(d)  - B(d)|$$
Here $\widehat{F}_X$ denotes the empirical cumulative distribution function for a set of values $X$, i.e.\ $\widehat{F}_X(d)$ is the empirically estimated probability that $x \le d$ for $x$ randomly sampled from $X$. 
$T$ corresponds to a Kolmogorov-Smirnov statistic between two distributions, in our case the \emph{foreground} and \emph{background} distribution of \emph{index-distances}. For IID data, these two distributions should be identical.

\subsection{Permutation Testing}
While there exists analytical methods to assess the significance of Kolmorogov-Smirnov statistics, these rely on asymptotics that require IID observations, whereas our ``observations'' here are index-distances between pairs of datapoints which are certainly not IID. Here we instead simply rely \emph{permutation testing} to obtain a p-value for the significance of our test statistic $T$ \cite{permutation}. For $P$ permutations, we permute the order of dataset $D$ and compute another test statistic for each permuted dataset, obtaining a collection of $P$ statistics distributed under our null hypothesis where the order of the data does not matter.

While permutation testing can be sometimes require intensive computation, we accelerate this process in multiple ways. First we only compute relatively few permutations (only 25 permutations were used for each result in this paper), and use kernel density estimation to extrapolate the null distribution from the $P$ permuted test statistics. Also done by \citet{mueller2018modeling}, such kernel density smoothing produces a smoother spectrum of possible p-values from a limited number of permutations. 
Additionally, each permutation is quite efficient because we don't need to recompute the kNN graph or background distribution $B(d)$.
For large datasets, the entire test can be done via a subset of the possible pairs from the dataset. In all results presented in this paper, we only consider at most $k$ pairings with each datapoint to avoid quadratic runtime complexity. Further speedups can be achieved by using approximate nearest neighbor algorithms to construct the initial kNN graph \cite{approximate}.

\section{Results}

\subsection{Comparison to other methods}

Using simulated datasets, we first compare our approach against alternative baseline methods to detect certain violations of the IID assumption. The first baseline is a simple \textbf{auto-correlation} measure. We use the Ljung-Box test \cite{ljungbox} on each dimension of numerical input data and average the p-values from this test across dimensions and lags to produce a single measure for a multivariate dataset. 

A second baseline method is based on \textbf{PCA reconstruction error}, a popular way to detect distribution drift \cite{nannyml}. This method computes the PCA reconstruction error over a series of subsets of data with respect to a reference subset. High error indicates the existence of data drift. This baseline method uses the same permutation testing framework as our kNN approach in order to convert PCA reconstruction errors into p-values. When working with 2-dimensional data (in our simulations), we use only the first principal component to compute reconstruction error.

When applying our kNN method, we always use the following default parameter settings: $k=10$ and 25 permutations and Euclidean distance measure. We compare the performance of these three methods on a few simulated 2-dimensional datasets (depicted in \cref{fig:toydata}) each with 1000 samples. The non-IID datasets stem from the following scenarios:

\begin{enumerate}
    \item \textbf{Gradual mean shift.} Data is drawn from a mixture of 10 Gaussians with means drawn uniformly from $[0, 10] \times [0, 10]$ and standard deviations drawn uniformly from $[0, 1]$. Each time a datapoint is drawn, the means increase linearly such that the mean of each Gaussian has shifted by 2 in each dimension by the end of data collection. This setting represents a classic example of distribution drift studied in the literature \cite{driftreview}.

    \item \textbf{Variance changepoint.} Data is drawn from the same mixture of Gaussians described in the previous paragraph. After the first half of the dataset has been collected, the standard deviation of each Gaussian in the underlying mixture distribution is multiplied by a factor of 1.5 after which the remaining half of the samples are collected.

    \item \textbf{Dependent but marginally identical in distribution.} In this scenario, we sample three points from a standard two-dimensional normal distribution. Then, each of the following datapoints, $x_i$ is sampled in an auto-regressive fashion from: $\mathcal{N}(\alpha_1 x_{i - 1} + \alpha_2 x_{i-2} + \alpha_3 x_{i-3}, 1)$ where $\alpha_1 + \alpha_2 + \alpha_3 = 0$. In this case, every datapoint is sampled from the same marginal distribution (identically distributed), but their values are not independent. For this demonstration, we use $\alpha_1 = 0.5, \alpha_2 = 0.4, \alpha_3 = -0.9$.
\end{enumerate}

\paragraph{Results of benchmark}
\cref{fig:baselines} shows the results of each method applied in each scenario, as well as a similar IID dataset created by randomly shuffling data from the scenario. Repeatedly running each method on datasets from each scenario produces a distribution of p-values for both non-IID and IID (shuffled) data. It is evident that only our kNN method reliably distinguishes between non-IID and IID data in every scenario, with a clear separation between p-values from the two settings. 

The auto-correlation baseline struggles in the \emph{variance changepoint} scenario in which virtually no temporal dependence exists on small timescales other than near the changepoint itself. In the final \emph{dependent data}  scenario in which the temporal dependence is the most salient, autocorrelation excels, but our kNN approach also reliably detects this form of IID violation. PCA Reconstruction fares poorly in many scenarios, producing p-value distributions between non-IID and IID runs that are not clearly distinguished except in the last scenario in which it performs better but still lags the other two methods. Only our kNN method demonstrates sensitivity to both data drift and interactions between datapoints' values without drift.

\begin{figure*}[tb]
    \centering
    \subfigure[Sorted by class]{\label{fig:sorted}\includegraphics[width=0.58\textwidth]{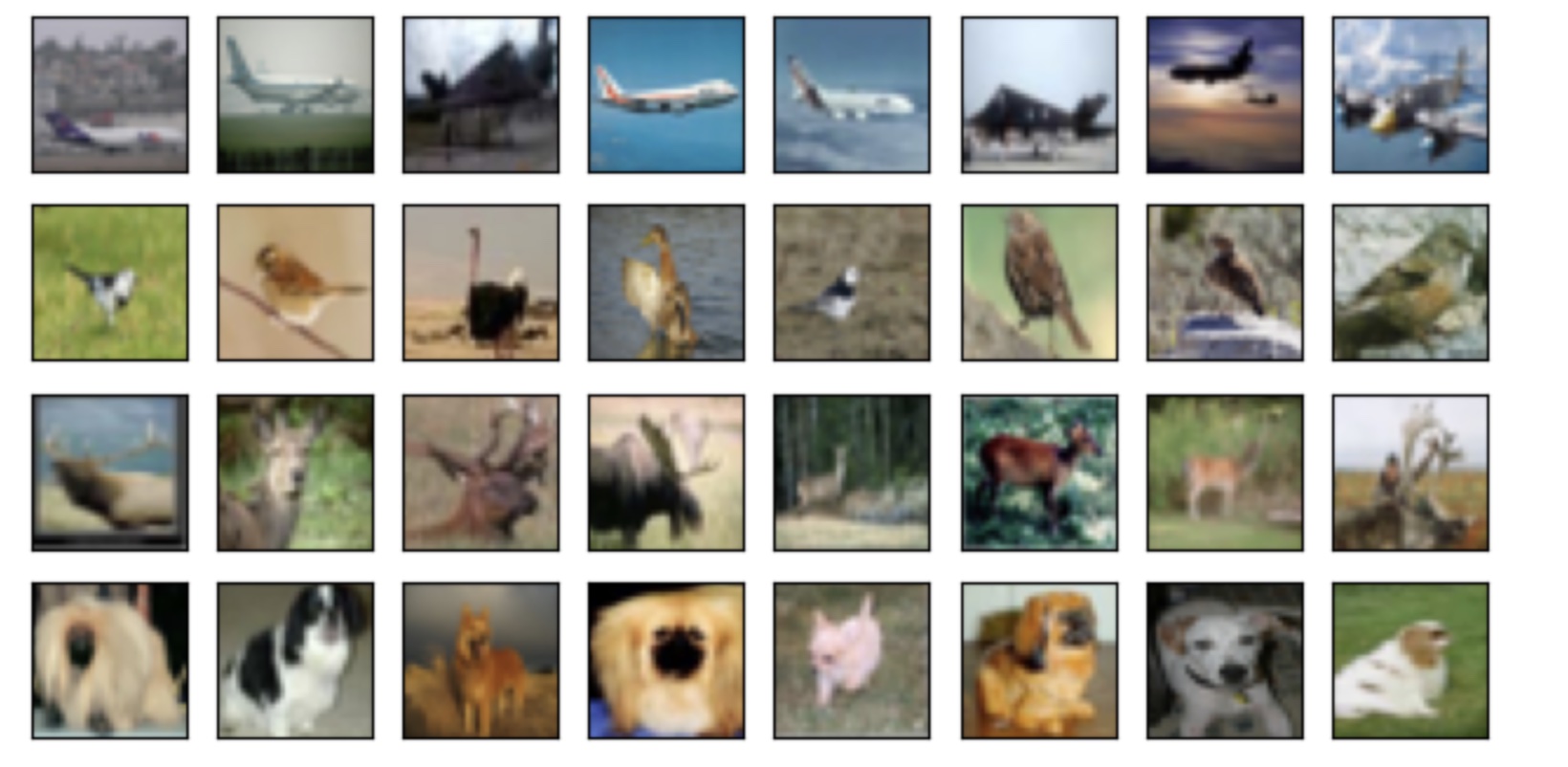}}
    \subfigure[Class distribution drift]{\label{fig:drift}\includegraphics[width=0.4\textwidth]{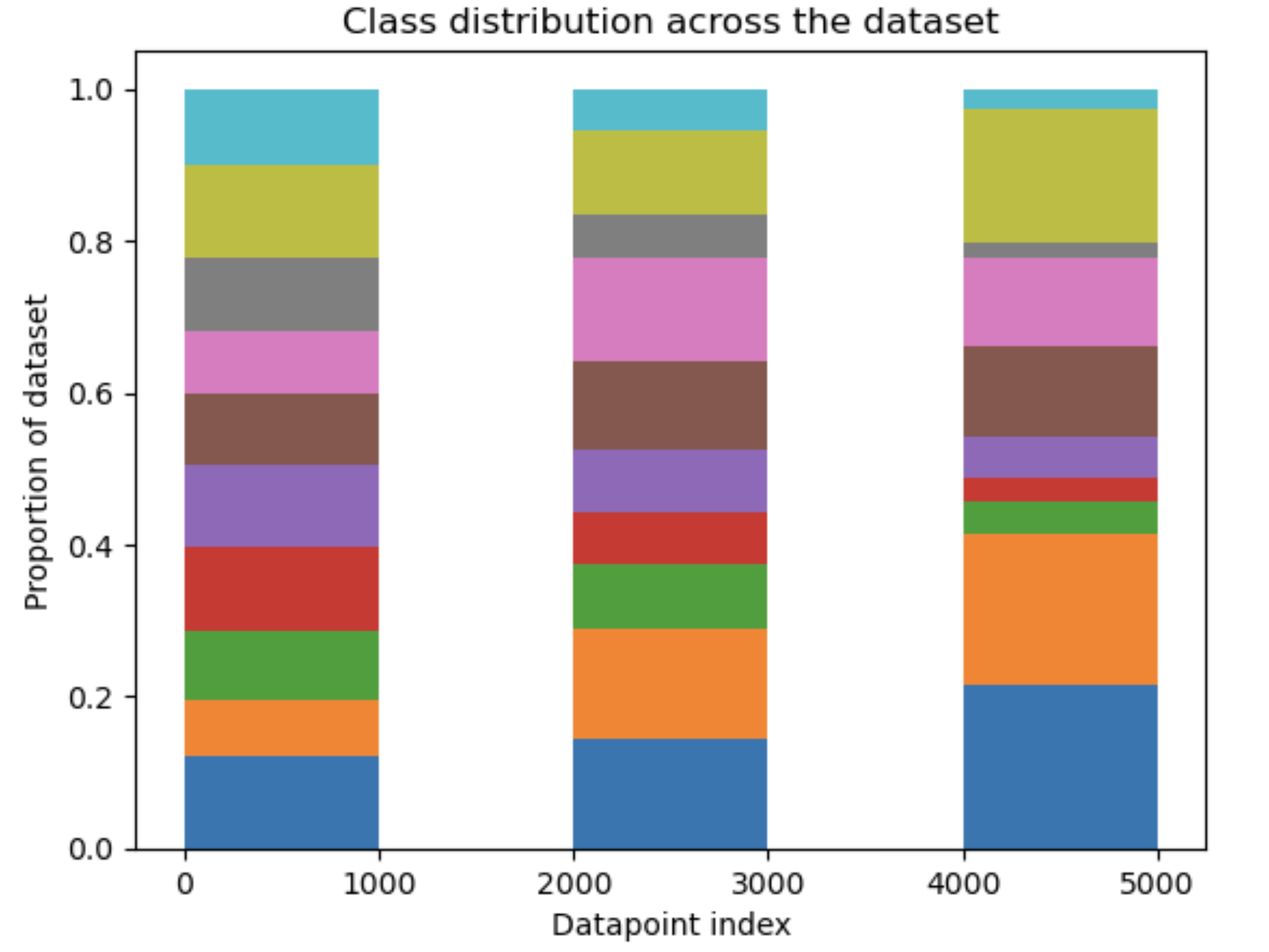}}
    \subfigure[Contiguous class subset]{\label{fig:contiguous}\includegraphics[width=0.58\textwidth]{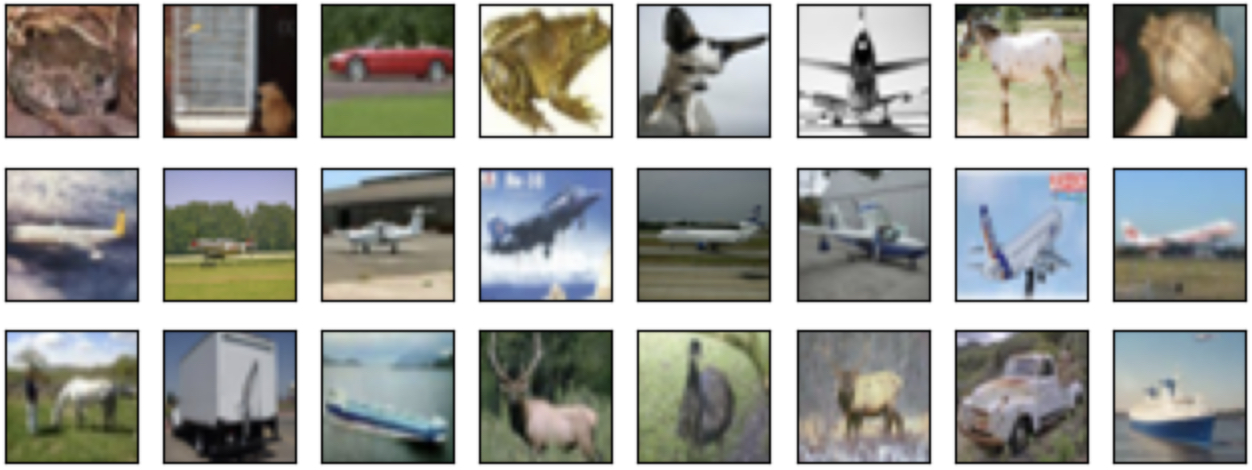}}
    \caption{Depicting our various non-IID scenarios using CIFAR-10. \textbf{(a)} Images are sorted by class. Each row of the example grid contains images of the same class which differs from row to row. \textbf{(b)} The distribution of classes drifts over time. We plot the class distributions of 3 sections of the data in the first, second and last thirds of the dataset. \textbf{(c)} The data contains a contiguous subset of images belonging to the same class. In this image grid, the first and last row are images drawn randomly from the whole dataset, but the middle row contains only images of random airplanes.}
    \label{fig:cifar10}
\end{figure*}
\begin{figure*}[h!]
    \centering
    \includegraphics[scale=0.55]{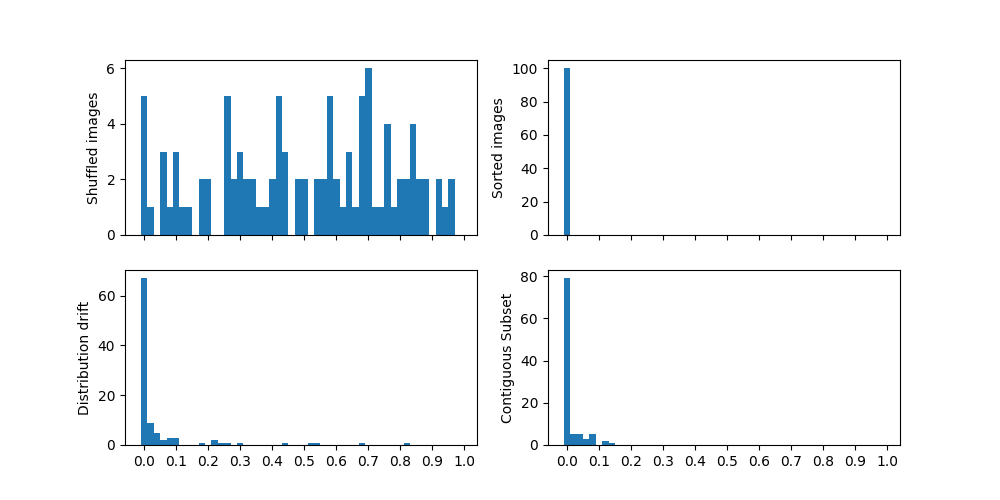}
    \caption{Results of our kNN method on the four image data scenarios described previously. Shown are histograms of p-values computed from 50 replicate runs of our method in each scenario, using default parameters ($k=10$, 25 permutations).}
    \label{fig:cifar_results}
\end{figure*}

\subsection{Image data scenarios}

We subsequently study how our kNN method works on a handful of non-IID subsets of the CIFAR-10 image classification dataset \cite{cifar10}. These showcase the generality of our methodology across multiple axes: it is able to detect many types of violations commonly expected in real-world data and directly applicable to complex image data. For images, we extract embeddings from an pretrained ResNet-50 model \cite{he2016deep,rw2019timm}, and then use the cosine distance between them to form the kNN graph. The alternative baseline methods fared  poorly in all image data scenarios we considered and are thus not presented here. In contrast, our method remains  effective and can be similarly applied to other data types with semantic model embeddings such as: text, audio, etc. Depicted in Figure \ref{fig:cifar10}, here are the image data scenarios we study:
\begin{enumerate}
\item \textbf{IID data.} First, we draw 10,000 random images from CIFAR10,  shuffle them, and apply our kNN method to ensure that it correctly identifies when data is, in fact, drawn IID at random from CIFAR-10.
\item \textbf{Data sorted by underlying class}. In the next scenario, we sort this 10,000 image subset by class so all images of a given class appear contiguously in the dataset. Note the class labels are \emph{not included} in the dataset to which we apply our method, solely the images themselves. 
This is an extreme example of non-IID data, where the underlying sampling distribution is subject to multiple sharp changepoints. It entails an easy setting that any non-IID detection algorithm ought to detect. 
\item \textbf{Distribution drift}. Our next experiment considers distribution drift. We create a 5,000 image subset which simulates this by randomly sampling images from CIFAR-10 from a probability distribution over their classes that is slowly evolving. At first, images are sampled from any object class with equal probability. Then, we gradually change the sampling weights to be less symmetric such that some classes are far more prevalent than others in the last third of the data.
\item \textbf{Contiguous non-IID subset in the middle of dataset}. 
In certain datasets, most of the data were collected in an IID manner, but something happened in the midst of data collection leading to a cluster of non-IID data lurking the broader dataset. For instance, accidentally including multiple frames from the same video in a web-scraped image dataset, or individuals from the same community in a survey dataset. Our next experiment studies such settings. 

We draw 2,500 images IID from CIFAR-10, except that we insert a small contiguous subset of 250 images that are drawn all from the same class halfway through the dataset (here we arbitrarily choose the \emph{airplane} class). \looseness=-1
\end{enumerate}

\paragraph{Results for image data}
\cref{fig:cifar_results} shows the results of our method in each of these scenarios, after many replicate runs of the method in each scenario.
Our method approach consistently detects non-IID sampling in each of the 3 non-IID scenarios with low p-values in almost every replicate run. In contrast, the p-values appear uniformly distributed for the IID dataset where images are randomly drawn from CIFAR-10, as expected under the null hypothesis (and guaranteed for permutation testing).

\section{Scoring individual datapoints to spot trends}

Thus far, we only considered producing a single p-value to summarize whether the collection order of a dataset appears to matter or not. This section presents a simple technique to gain more insights about a dataset that produced a low p-value. Specifically, we score every datapoint in the dataset and plot these scores against the index of the datapoints to see if any trends emerge.

Our score is obtained for each datapoint $x_i$ by computing the same sort of Kolmogorov-Smirnov statistic $T$ used in our overall hypothesis test, but this time restricting the two distributions being compared, $\widehat{F}$ and $B$, to only consider the portions of the foreground/background distribution in which $x_i$ is involved. That is we only consider pairs of datapoints in which the first element is $x_i$ itself (neighbors of $x_i$ in the kNN graph for the foreground distribution, random other datapoints paired with $x_i$ for the background distribution). For ease of visualization, we map the resulting per-datapoint statistic to a score between 0 and 1, such that scores near 0 indicate a datapoint for which the \emph{index-distance} distributions to its neighbors and non-neighbors are significantly different. \looseness=-1

\begin{figure}[h!]
    \centering
    \subfigure[IID data]{\label{fig:sorted}\includegraphics[width=0.37\textwidth]{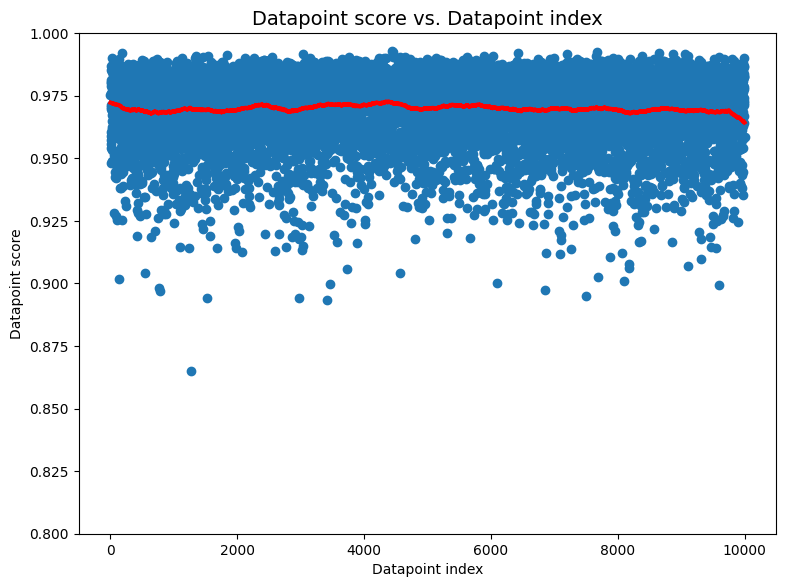}}
    \subfigure[Class distribution drift]{\label{fig:drift}\includegraphics[width=0.37\textwidth]{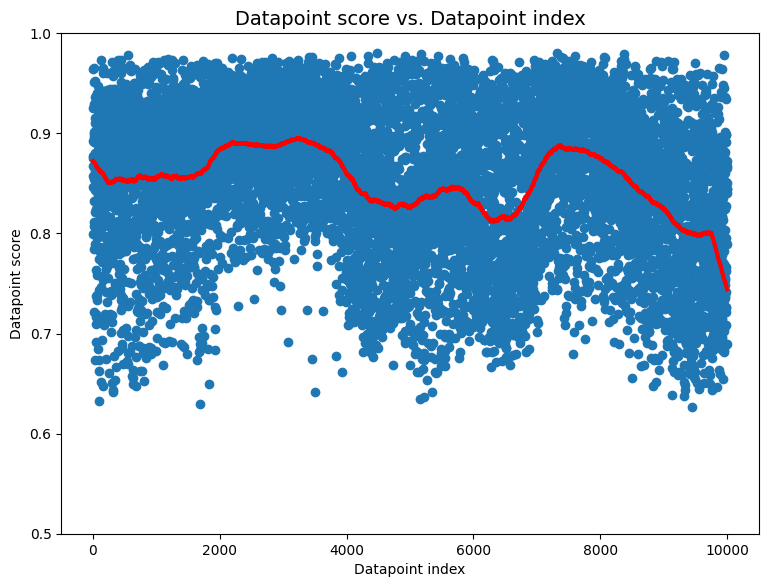}}
    \subfigure[Contiguous class subset]{\label{fig:contiguous}\includegraphics[width=0.37\textwidth]{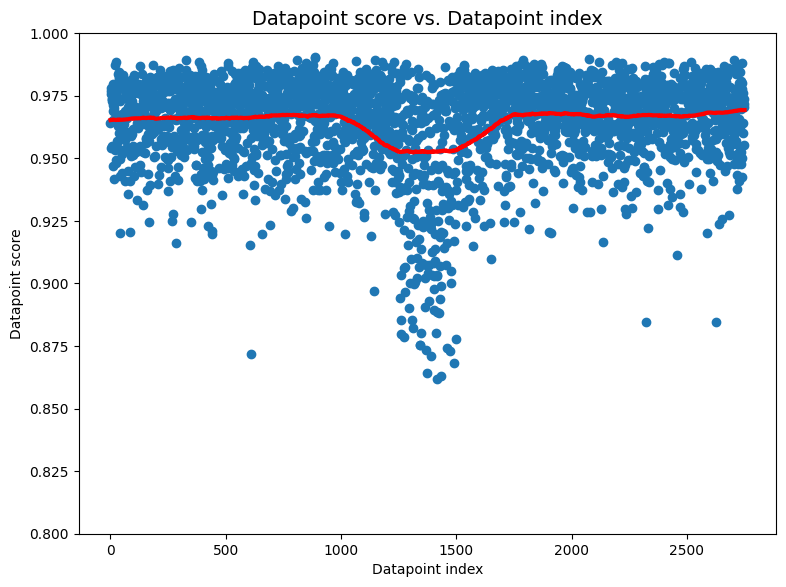}}
    \caption{Per datapoint scores plotted against the datapoint index for various scenarios involving the CIFAR-10 image data. Shown in \textcolor{red}{red} is a low-pass filter to smooth the scores and highlight their trend. The scenarios are as follows: 
\textbf{(a)} Images are randomly drawn from CIFAR-10 (IID). \textbf{(b)} The \emph{data sorted by underlying class} scenario depicted in Figure \ref{fig:sorted} in which the underlying latent class changes every 1000 images. \textbf{(c)} The \emph{contiguous non-IID subset in the middle of dataset} scenario depicted in Figure \ref{fig:contiguous}. Here all images between the 1250th and 1500th in the dataset happen to be from the airplane class, while other images are randomly drawn from CIFAR-10.}
    \label{fig:datapointscores}
\end{figure}

\cref{fig:datapointscores} shows what these scores look like for our previously described image data scenarios. The results display no trend in these per-datapoint scores when the dataset is IID. However informative trends can be seen for scenarios where the dataset is not IID and p-values from our kNN approach are low. In the \emph{data sorted by underlying class} scenario, we see clear peaks and valleys in these scores whose span corresponds to the number of images sampled from a particular class before drawing images from another class. 
In the \emph{contiguous non-IID subset in the middle of dataset} scenario, we see a clear valley in these scores whose location corresponds to the non-IID data (images all from the airplane class) that happens to be present in an otherwise IID dataset. \looseness=-1

In practice, a data analyst who receives a low overall p-value for their dataset can zoom in the region of the dataset highlighted by these scores. Understanding how this region differs from other regions of the dataset with different score behavior may reveal important insights about data collection mishaps or important trends to account for during modeling.

\section{Discussion}
This paper presents a simple/scalable approach to detect certain types of common violations of the IID assumption. We statistically test whether or not datapoints that are closer in the \emph{data ordering} (i.e.\ collection time) also tend to have more similar \emph{feature values}.
Our approach exhibits high power to detect many forms of such issues, and can be applied to many different data types, as long as suitable similarity measure between two datapoints' features can be defined. Our kNN method's  capabilities extend beyond drift detection to properly catching settings where nearly adjacent datapoints are positively interacting (lack of independence).

This generality is a strength, but may also be seen as a limitation. A more specialized algorithm may capture more information about \emph{how} a non-IID trend presents itself, however many kinds of problematic data will go undetected as observed in our empirical results. Our work represents a step toward developing automated data investigation methods that algorithmically detect fundamental problems in any given dataset.
As more data analysis is conducted by non-experts, such automated data checks will become increasingly vital to ensure reliable inferences are produced.

\vspace*{2em}
\balance{}
\bibliographystyle{icml2023}
\bibliography{main}
\balance{}

\end{document}